\documentclass[letterpaper, 10 pt, conference]{ieeeconf}  

\usepackage[style=ieee, doi=false, isbn=false, backend=bibtex]{biblatex}
\usepackage[hidelinks]{hyperref}

\usepackage{shellesc}

\usepackage[super]{nth} 

\usepackage{amsmath}%
\usepackage{amsfonts}%
\usepackage{amssymb}%
\usepackage{mathtools}%
\usepackage{mathrsfs} 

\usepackage{tabularx}
\usepackage{booktabs}
\usepackage[table]{xcolor}
\newcolumntype{Y}{>{\centering\arraybackslash}X}
\usepackage[exponent-product = \cdot,
			output-complex-root = j, 
			separate-uncertainty = true,
			output-product = \cdot,
			arc-separator = \,,
			product-units = brackets-power]{siunitx}
\usepackage{subcaption}

\usepackage{tikz,pgfplots}
   
\def\MPC{MPC}

\def\OCP{OCP}
\def\OCPc{OCP}

\newcommand*{\mtext}[1]{\text{\normalfont #1}} 

\newcommand*{\x}{\ensuremath{\mathbf{x}}}%
\newcommand*{\xd}[1]{\ensuremath{\x_{#1}}}%
\newcommand*{\xc}[1]{\ensuremath{\x(#1)}}%
\newcommand*{\xcdot}[1]{\ensuremath{\dot{\x}(#1)}}%
\newcommand*{\dx}{\ensuremath{\Delta \x}}%

\renewcommand*{\u}{\ensuremath{\mathbf{u}}}%
\newcommand*{\uc}[1]{\ensuremath{\u(#1)}}%
\newcommand*{\ucdot}[1]{\ensuremath{\dot{\u}(#1)}}%

\newcommand*{\xs}{\ensuremath{\x_{\mtext{s}}}}%
\newcommand*{\ts}{\ensuremath{t_{\mtext{s}}}}%

\newcommand*{\f}{\ensuremath{\mathbf{f}}}

\newcommand*{\xu}{\ensuremath{\x}}
\newcommand*{\xuc}[1]{\ensuremath{\xu(#1)}}%
\newcommand*{\xud}[1]{\ensuremath{\x_{#1}}}%
\newcommand*{\uud}[1]{\ensuremath{\u_{#1}}}%
\newcommand*{\uucstar}[1]{\ensuremath{\u^*(#1)}}%

\newcommand*{\vardt}{\ensuremath{\Delta t}}
\newcommand*{\vardtstar}{\ensuremath{\Delta t^*}}
\newcommand*{\vardtref}{\ensuremath{\vardt_\mtext{s}}}
\newcommand*{\vardthyst}{\ensuremath{\vardt_\epsilon}}
\newcommand*{\vardtmax}{\ensuremath{\vardt_\mtext{max}}}

\newcommand*{\vardtmin}{\ensuremath{\vardt_\mtext{min}}}

\newcommand*{\Nmin}{\ensuremath{N_\mtext{min}}}%

\newcommand*{\dmin}{\ensuremath{d_\mtext{min}}}%

\newcommand*{\xf}{\ensuremath{\x_\mtext{f}}}%
\newcommand*{\tf}{\ensuremath{t_\mtext{f}}}

\newcommand*{\mulaw}{\ensuremath{\mu}}%

\newcommand*{\nullvec}{\ensuremath{\mathbf{0}}}%

\newcommand*{\Jf}{\ensuremath{J_\mtext{f}}}

\newcommand*{\Qf}{\ensuremath{\mathbf{Q}_\mtext{f}}}
\newcommand*{\Q}{\ensuremath{\mathbf{Q}}}
\newcommand*{\R}{\ensuremath{\mathbf{R}}}

\newcommand*{\realpos}{\ensuremath{\mathbb{R}^+}}%
\newcommand*{\realposzero}{\ensuremath{\mathbb{R}^+_0}}%
\newcommand*{\naturalzero}{\ensuremath{\mathbb{N}_0}}%
\newcommand*{\naturalpos}{\ensuremath{\mathbb{N}}}

\newcommand*{\xset}{\ensuremath{\mathcal{X}}}%
\newcommand*{\rxset}{\ensuremath{\mathbb{X}}}%
\newcommand*{\uset}{\ensuremath{\mathcal{U}}}%
\newcommand*{\ruset}{\ensuremath{\mathbb{U}}}%
\newcommand*{\rudset}{\ensuremath{\mathbb{U}_\mtext{d}}}%

\newcommand*{\rxoset}{\ensuremath{\mathbb{X}_\mtext{o}}}%
\newcommand*{\rxiset}{\ensuremath{\mathbb{X}_\mtext{i}}}%

\newcommand*{\xfset}{\ensuremath{\mathbb{X}_\mtext{f}}}%

\newcommand*{\optimparamglobal}[1]{\ensuremath{\mathcal{P}_\mtext{global}}}

\newcommand*{\z}{\ensuremath{\mathbf{z}}}%
\newcommand*{\zstar}{\ensuremath{\mathbf{z}^*}}%
\newcommand*{\deltaz}{\ensuremath{\Delta \mathbf{z}}}%
\DeclareMathOperator{\nablaop}{\nabla}
\newcommand*{\gradwrt}[2]{\ensuremath{\nablaop_{\kern -0.2em #2} #1}} 
\DeclareMathOperator{\Djac}{D}

\newcommand*{\jacobwrt}[2]{\ensuremath{\Djac_{#2} #1}}

\newcommand*{\hessianwrt}[2]{\ensuremath{\nablaop^2_{\kern -0.2em #2} #1}}

\newcommand{\dt}{\text{d}t}

\newcommand{\transpose}{\intercal}

\newcommand{\fundef}[3]{#1\,{:}\,#2\,{\mapsto}\,#3}
\newcommand{\mdef}{\vcentcolon=}

\newcommand{\reffig}[1]{Fig.~\ref{#1}}

\newcommand{\refsec}[1]{Section~\ref{#1}}

\newcommand{\refexample}[1]{Example~\ref{#1}}

\newcommand{\reffigc}[1]{Fig.~\ref{#1}}
\newcommand{\reftabc}[1]{Table~\ref{#1}}

\newcommand{\refsecc}[1]{Section~\ref{#1}}

\newtheorem{rem}{Remark}

\newtheorem{example}{Example}

\usepgfplotslibrary{patchplots}
\usetikzlibrary{positioning,automata,through,math,calc,plotmarks,shapes,arrows,arrows.meta,quotes,decorations.markings,shapes.misc,shapes,backgrounds,patterns,angles,babel,intersections,decorations,spy,decorations.pathreplacing, shadows,shadings} 
\pgfplotsset{compat=newest}
\pgfplotsset{plot coordinates/math parser=false}

\pgfplotsset{/pgf/number format/.cd,1000 sep={}} 

\newlength\figureheight
\newlength\figurewidth

\tikzstyle{every picture}+=[remember picture]
\tikzstyle{arrow} = [->,>=stealth']
\tikzstyle{arrowreverse} = [<-,>=stealth']
\tikzstyle{arrowbidir} = [<->,>=stealth']

\tikzset{thin/.style ={line width= 0.2mm}}
\tikzset{thick/.style ={line width= 0.4mm}}
\tikzset{very thick/.style ={line width= 0.5mm}}

\pgfplotsset{tick style={black}} 

\pgfplotsset{every axis plot post/.append style={line join=round}}

\tikzset{every picture/.style={font issue=\small},
	font issue/.style={execute at begin picture={#1\selectfont}}
}
\tikzset{fontscale/.style = {font=\small}
}

\pgfplotsset{every axis/.append style={
		axis lines = left 
	}
}

\tikzset{%
	partially dashed/.style={
		decoration={show path construction, 
			lineto code={
				\draw[] (\tikzinputsegmentfirst) --($(\tikzinputsegmentfirst)!#1!(\tikzinputsegmentlast)$);,
				\draw[dashed, dash phase=3pt,-latex] ($(\tikzinputsegmentfirst)!#1!(\tikzinputsegmentlast)$)--(\tikzinputsegmentlast);,
			}
		},
		decorate
	},
}

\makeatletter
\pgfplotsset{ 
	every axis x label/.append style={
		alias=current axis xlabel,
	},
	legend pos/outer south/.style={
		/pgfplots/legend style={
			at={%
				(%
				\@ifundefined{pgf@sh@ns@current axis xlabel}%
				{xticklabel cs:0.5}%
				{current axis xlabel.south}%
				)%
			},
			anchor=north,
			legend columns=3,
			font=\small,
			fill=none,
			/tikz/every even column/.append style={column sep=10pt, font=\small} 
		}
	},
	legend pos/outer north/.style={
		/pgfplots/legend style={
			at={(\figurewidth/2,\figureheight+0.2cm)},
			draw=none, 
			anchor=south,
			legend columns=3,
			font=\scriptsize,
			fill=none,
			/tikz/every even column/.append style={column sep=10pt, font=\scriptsize} 
		}
	},
	legend pos/outer north2/.style={
		/pgfplots/legend style={
			at={(\figurewidth/2,\figureheight+0.5cm)},
			draw=none, 
			anchor=south,
			legend columns=3,
			font=\scriptsize,
			fill=none,
			/tikz/every even column/.append style={column sep=5pt, font=\scriptsize} 
		}
	}
}
\makeatother

\newlength\customshift    

\usepackage{paralist}

\makeatletter
\DeclareBibliographyOption[boolean]{dashed}[true]{%
	\ifstrequal{#1}{true}
	{\ExecuteBibliographyOptions{pagetracker}%
		\renewbibmacro*{bbx:savehash}{\savefield{fullhash}{\bbx@lasthash}}}
	{\renewbibmacro*{bbx:savehash}{}}}
\makeatother
\IEEEoverridecommandlockouts                              

\title{\LARGE \bf
Online Motion Planning based on Nonlinear Model Predictive Control with Non-Euclidean Rotation Groups
}

\author{Christoph R\"osmann, Artemi Makarow and Torsten Bertram
	\thanks{The authors are with the Institute of Control Theory and Systems Engineering, TU Dortmund University, 44227 Dortmund, Germany
		{\tt\small \{forename.surname\}@tu-dortmund.de}}%
}

\begin{document}

\maketitle
\thispagestyle{empty}
\pagestyle{empty}

\begin{abstract}
This paper proposes a novel online motion planning approach to robot navigation based on nonlinear model predictive control.
Common approaches rely on pure Euclidean optimization parameters.
In robot navigation, however, state spaces often include rotational components which span over non-Euclidean rotation groups.
The proposed approach applies nonlinear increment and difference operators in the entire optimization scheme to explicitly consider these groups. 
Realizations include but are not limited to quadratic form and time-optimal objectives.
A complex parking scenario for the kinematic bicycle model demonstrates the effectiveness and practical relevance of the approach.
In case of simpler robots (e.g. differential drive), a comparative analysis in a hierarchical planning setting reveals comparable computation times and performance.
The approach is available in a modular and highly configurable open-source C++ software framework. 
\end{abstract}



\section{INTRODUCTION}

In the context of robotics and autonomous driving, online trajectory planning, usually as part of a hierarchical planning architecture, is still an essential part of research  to meet the requirements of navigation in increasingly complex and highly dynamic environments.
In contrast to classical path planning approaches~\cite{lavalle2006,lamiraux2004}, trajectory planning takes the temporal profile into account and enables improved performance as well as explicit compliance with kinodynamic and dynamic constraints.
An established method is the extension of the well-known elastic band (EB) path planning approach~\cite{quinlan1995} to trajectory deformation in~\cite{delsart2008}.
Lau et al. present a method that represents trajectories by Bézier splines and adheres to kinodynamic constraints of non-holonomic robots~\cite{lau2009}.
Not only for mobile robots, but especially for integrator dynamics, online planning algorithms based on covariant or stochastic gradient descent and obstacle potentials are provided in~\cite{ratliff2009,kalakrishnan2011}.
Delsart et al. extend the EB approach to the deformation of trajectories rather than paths~\cite{delsart2008}.
Lau et al.~\cite{lau2009} optimize trajectories represented by splines according to kinodynamic constraints of the robot.

\begin{figure}[t]
	\begin{tikzpicture}[font=\footnotesize]
	\tikzstyle{box1}=[draw, rectangle,minimum width=3.7cm, minimum height=0.8cm, align=center, rounded corners=1ex]
	\tikzstyle{box2}=[box1, minimum width=3.9cm]
	\tikzstyle{colored}=[fill=gray!15]
	
	\node[box1] (ocp) at (0,0) {\textbf{Optimal control problem}\\ \scriptsize Optimize w.r.t. controls (+states/time)};
	
	\node[box1, xshift=-2.3cm, yshift=-1.6cm] (mpc_nlp) at (ocp) {\textbf{Nonlinear program}\\ \scriptsize (or nonlinear system of equations)};
	\node[box1, colored, yshift=-1.27cm] (mpc) at (mpc_nlp) {\textbf{Model predictive controller}};
	
	\node[box2, xshift=2.3cm, yshift=-1.6cm] (flat_sys) at (ocp) {\textbf{Flat system model}\\ \scriptsize Optimize w.r.t. flat output (+time)};
	\node[box2, yshift=-1.27cm] (flat_nlp) at (flat_sys) {\textbf{Nonlinear program}};
	\node[box2, yshift=-1.27cm] (soft_constr) at (flat_nlp) {\textbf{Soft-constraint approximation}\\ \scriptsize Unconstrained optimization problem};
	\node[box2, colored, yshift=-1.27cm] (teb) at (soft_constr) {\textbf{Timed-Elastic-Band approach}\\ \scriptsize Time-optimal, \scriptsize finite-diff. collocation};
	
	\draw [-latex] (ocp) -- (mpc_nlp) node [pos=0.5,align=center, left=0.45cm] {Direct methods\\(or indirect methods)};
	\draw [-latex] (mpc_nlp) -- (mpc) node [pos=0.5,align=center, left] {State feedback};
	\draw [-latex] (ocp) -- (flat_sys) node [pos=0.5,align=center, right=0.45cm] {Conversion of\\cost \& constraints};
	\draw [-latex] (flat_sys) -- (flat_nlp) node [pos=0.5,align=center, right] {Direct methods};
	\draw [-latex] (flat_nlp) -- (soft_constr) node [pos=0.5,align=center, right] {Penalty methods};
	\draw [-latex] (soft_constr) -- (teb) node [pos=0.5,align=center, right] {State feedback};
	\end{tikzpicture}
	\caption{From optimal control to online motion planning}
	\label{fig:ocp_to_motion_planning}
	\vspace{-\baselineskip}
\end{figure}
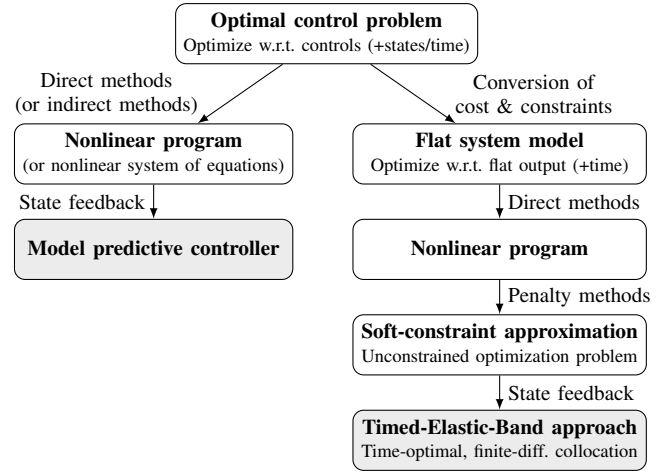

Many optimization based approaches can be derived from a generic optimal control formulation~\cite{gulati2011, bonalli2019} and interpreted with state feedback as a variant of predictive control. 
Predictive controllers repeatedly solve an optimal control problem (OCP) during runtime while commanding the first action of the optimal control trajectory to the system in each closed-loop step.
\reffigc{fig:ocp_to_motion_planning} shows two examples for such derivations. The left side shows what is commonly known as model predictive control (MPC) of continuous-time systems.
Direct methods discretize the OCP to obtain a nonlinear program (NLP) for which many efficient solving techniques exist~\cite{betts2010_book}.
State feedback then completes the approach to a model predictive controller. Note that the controller type depends on the OCP formulation,
e.g., receding horizon MPC and variable horizon MPC.  
Since the computational burden is large, many approaches restrict the solution space or approximate the original problem.
For example the dynamic window approach (DWA)~\cite{fox1997} can be interpreted as a special receding horizon MPC scheme~\cite{ogren2005}.
The search space is restricted to a (collision-free) constant control input for the complete horizon. 
The required computation times are low but due to the less degrees of freedom in control, the approach is suboptimal and cannot predict motion reversals such that control of car-like robots is rather limited.
Another approach is the Timed-Elastic-Band (TEB) approach, which was originally derived as a scalarized multi-objective least squares optimization~\cite{roesmann2012}.
It can also be derived from a time-optimal control problem as shown in the right of \reffig{fig:ocp_to_motion_planning}.
Hereby, the robot kinematics are expressed geometrically similar to a flat system model in the pose space and a nonlinear program is obtained by applying direct transcription, in particular finite-difference collocation.
The nonlinear program is then transformed to an unconstrained least-squares problem by applying quadratic penalty functions~\cite{roesmann2017}.
In~\cite{roesmann2017_iros}, the TEB is extended to car-like robots, however, the flat system model limits the definition of arbitrary constraints, e.g. limits on the steering rate are not included yet.
Recently, an extension and reformulation in terms of Lie groups is proposed in~\cite{deray2019}.

Since computational resources have increased and algorithms have become more efficient, the interest in full MPC-based approaches has been growing.
Path following control with recursive feasibility and stability properties is provided in~\cite{faulwasser2016_tac}. 
Zhang et al. generate collision-free trajectories and considers general obstacles and that can be represented as the union of convex sets~\cite{zhang2017optimizationbased}. 
A recent approach based on convex inner approximations provides feasible and collision-free solutions in a few solver iterations~\cite{schoels2019nmpc}. 

Motion planning for robot navigation usually includes non-Euclidean rotational components, i.e. the robot's heading. 
The contribution of this paper is as follows: We propose an MPC-based motion planning scheme that differs from others by the explicit consideration of orientation groups during optimization.
To our best knowledge, available MPC-based approaches are tailored for Euclidean state spaces. 
Our MPC formulation is versatile and includes many common realizations including receding horizon quadratic form and shrinking horizon time-optimal objectives.
We further provide a generic, highly customizable and easily extendable C++ software framework with Robot Operating System (ROS) integration (\textit{mpc\_local\_planner}).

The paper is organized as follows:
\refsecc{sec:prob_description} provides the formal problem description of the approach and \refsec{sec:realization} the numerical realization.
Section~\ref{sec:evaluation} first evaluates the proposed approach with a kinematic bicycle model
and then conducts a comparative analysis for a common navigation scenario.
Finally, section~\ref{sec:conclusions} concludes the work.


\section{MOTION PLANNING PROBLEM DESCRIPTION}
\label{sec:prob_description}

This paper addresses robotic systems described by nonlinear and time-invariant differential equations with time $t \in \mathbb{R}$, state trajectory $\fundef{\x}{\mathbb{R}}{\xset}$ and control trajectory $\fundef{\u}{\mathbb{R}}{\uset}$:  
\begin{equation}
\xcdot{t} = \f\big(\xc{t},\uc{t}\big). 
\label{eq:system_dynamics_cont}
\end{equation}
The mapping $\fundef{\f}{\xset \times \uset}{\mathbb{R}^p}$ with $p=\dim(\xset)$ is continuous and Lipschitz in its first argument.
The state and control spaces, $\xset$ and $\uset$ respectively, do not have to be exclusively Euclidean spaces, as discussed later. 
System~\eqref{eq:system_dynamics_cont} is also subject to state and input constraint sets originating from internal robot constraints, but also from the dynamic environment in which the robot operates.
Therefore, state constraints $\xc{t} \in \rxset(t) \subseteq \xset$ and input constraints $\uc{t} \in \ruset\big( \xc{t}, t \big) \subseteq \uset$ are time-variant. 

\subsection{Local Planning via Optimal Control}

The planning task is to guide the robotic system~\eqref{eq:system_dynamics_cont} from $\xs \in \xset$ at time $\ts$ to an intermediate or ultimate goal set $\xfset(\ts) \subset \xset$ within time $\tf \in I$ while minimizing an objective function and adhering to constraints.
Hereby, $I \subseteq \mathbb{R}$ denotes a predefined time interval containing~$\ts$. 
With running cost $\fundef{\ell}{\xset \times \uset}{\realposzero}$ and terminal cost $\fundef{\Jf}{\xset}{\realposzero}$, the \OCP{} is given as follows:
\begin{gather}
	\underset{\uc{t},\, \xc{t},\, \tf}{\min} \quad \Jf\big(\xc{\tf}\big) + \int_{t=\ts}^{\tf} \ell\big(\xc{t}, \uc{t}\big)\,\dt
	\label{eq:ocp} \\
	\hspace{-6.8cm}\text{subject to} \nonumber \\ 
	\begin{align*}
&\xc{\ts} = \xs,   \quad \xcdot{t} = \f\big( \xc{t}, \uc{t} \big),  \quad \xc{\tf} \in \xfset(\ts), \\
&\xc{t} \in \rxset(t), \ \uc{t} \in \ruset\big(\xc{t}, t\big), \ \ucdot{t} \in \rudset, \ \tf \in I.
	\end{align*}
\end{gather}
State and control constraints are included as described before. 
In addition, the control derivative $\ucdot{t}$ is restricted to a given set $\rudset$.  
Note, the same result can be achieved by augmenting the state space with integrator dynamics.
However, in robot applications the system is often described by a kinematic model with velocities as input, and therefore such a constraint (i.e. acceleration limits) eliminates the need to increase the number of optimization parameters.
\OCPc{}~\eqref{eq:ocp} is generic at this point and includes the most common objectives like minimizing time or minimizing control error and effort (quadratic form). These are detailed in~\refsec{sec:realization}.
Note that potential fields for increasing the distances to obstacles or attraction terms for approaching waypoints may also be included in $\ell(\cdot)$.

\subsection{Feedback Control}

An MPC-based local planner provides control actions directly to the robot or to a cascaded low-level controller.
In practice, \eqref{eq:ocp} can only be solved at discrete time instances and hence the control law is defined according to the grid 
$t_0 < t_1 < \dotsc < t_n < \dotsc < \infty$ with \mbox{$n \in \naturalzero$} and $t_{n} \in \realposzero$.
The control law $\fundef{\mulaw}{\xset}{\uset}$ for $t \in [t_n, t_{n+1})$ is given by:
\begin{equation}
\mulaw\big(\xc{t}\big) \mdef \uucstar{t} \big\vert_{\xs=\xc{t_n}, t_s = t_n}.
\label{eq:mulaw}
\end{equation}
Hereby, $\uucstar{t}$ denotes the resulting optimal control trajectory from \OCP{}~\eqref{eq:ocp} with substitutions $\xs=\xc{t_n}$ and $t_s = t_n$.
The current state $\xc{t_n}$ is either directly measurable or estimated by a state observer.

\section{NUMERICAL REALIZATION}
\label{sec:realization}

This section transforms the generic motion planning problem into a numerically tractable realization.

\subsection{Alternative State Spaces}
\label{sec:alternative_ss}

Common MPC formulations usually assume that the state space is Euclidean (resp. the real \textit{n}-space). 
But especially in robotics, state spaces often include rotational components which span over the non-Euclidean rotation groups \textit{SO(2)} or \textit{SO(3)}.
A possible approach is to provide an over-parameterized formulation with constraints. 
However, local derivative-based optimization schemes apply local increments $\zstar = \z + \deltaz$ with parameter \mbox{$\z \in \mathbb{R}^M$} and increment $\deltaz \in \mathbb{R}^M$ over an $M$-dimensional Euclidean parameter space in each solver iteration~\cite{nocedal2006_book} without maintaining any of these constraints.
To overcome these difficulties, we adapt and extend the idea of alternative parameterizations from graph-based SLAM \cite{kuemmerle2011}, 
which we have already successfully applied in simplified form in the TEB implementation. 
The idea follows the observation that $\deltaz$ is usually a small perturbation around $\z$ and hence far from singularities. 
Therefore, $\deltaz$ represents a minimal representation computed in the local Euclidean surroundings of $\z$.
A nonlinear increment operator then converts and applies $\deltaz$ to the correct space.
Due to the limited scope of this paper, we restrict ourselves to 2D rotation groups \textit{SO(2)} which applies to most robot navigation scenarios.
In this case, even~$\z$ does not need to be over-parameterized but the nonlinear increment operator still maintains proper 2D rotations.

Consider a state $\x \in \xset$. An increment $\dx \in \mathbb{R}^p$ is then applied by the nonlinear increment operator $\fundef{\boxplus}{\xset \times \mathbb{R}^p}{\xset}$:
\begin{equation}
\x^* = \x \boxplus \dx.
\end{equation}
Hereby, $\mathbb{R}^p$ is a local Euclidean space.
To properly evaluate the system dynamics in the alternative state space, the difference $\xd{2} - \xd{1}$ for $\xd{1},\xd{2} \in \xset$ must be embedded into $\xset$. 
This is achieved by defining a nonlinear difference operator $\fundef{\boxminus}{\xset \times \xset}{\xset}$ such that $\delta \x = \xd{2} \boxminus \xd{1}$.
Note that no singularities occur for rotational components in \textit{SO(2)} and the operator defines the shortest angular distance between two states without discontinuities.
In this work we apply these operators only to states, but the extension to alternative control input spaces is done in exactly the same way.

\begin{example} \label{ex:se2}
	As an example consider the special Euclidean group~2, i.e. $\xset = \mtext{\textit{SE(2)}} = \mathbb{R}^2 \times \mtext{\textit{SO(2)}}$  that is defined in terms of a 2D translation in $\mathbb{R}^2$ and a rotational component.
	The state vector is defined as $\x = ( x, y, \theta )^\transpose \in \xset$ with $\theta \in [-\pi,\pi)$. Let $\operatorname{normAngle}(\varphi)$ define a function that normalizes an angle $\varphi\in\mathbb{R}$ to the interval $[-\pi,\pi)$.
	Then operator~$\boxplus$ is specified as:
	\begin{equation}
	\x \boxplus \dx \mdef \big(x+\Delta x, y + \Delta y, \operatorname{normAngle}(\theta + \Delta \theta)\big)^\transpose.
	\end{equation}
	Accordingly, the difference operator is: $\xd{2} \boxminus \xd{1} \mdef \xd{2} \boxplus -\xd{1}$.
\end{example}

\subsection{Direct Transcription}

This section applies direct transcription~\cite{betts2010_book} in combination with the previously defined increment and difference operators to convert \OCP{}~\eqref{eq:ocp} to an NLP.
The time interval $[\ts, \tf]$ of the planning horizon is now discretized according to the following grid:
$\ts = t_0 \leq t_1 \leq \dotsc \leq t_k \leq \dotsc \leq t_N = \tf$ with $t_k \in I$, $k=0,1,\dotsc,N$ and $N\in\naturalpos$. 
Note that index $k$ indicates that the context belongs to prediction rather than closed-loop control.
Furthermore, $\vardt_k = t_{k+1} - t_k$ denotes the time interval for an individual grid partition.
In this paper, direct transcription relies on a piecewise constant control trajectory with respect to the temporal grid, i.e.
$\uc{t} \mdef \uud{k} = \mtext{constant}$ for $t \in [t_k, t_k+\vardt_k)$ for $k=0,1,\dotsc,N-1$.
The states at grid points~$t_k$ are denoted as $\xuc{t_k} \mdef \xud{k}$ for $k=0,1,\dotsc,N$.

Several methods exist to discretize the boundary value problem in \OCP{}~\eqref{eq:ocp} induced by system~\eqref{eq:system_dynamics_cont}.
Established methods are multiple shooting and collocation~\cite{betts2010_book}. 
Whereas multiple shooting usually applies explicit integration schemes, collocation mainly refers to implicit schemes. 
In the following, we utilize low order collocation via finite-differences as they provide a reasonable trade-off between computational resources and accuracy
and they are directly suitable for the nonlinear difference operator $\boxminus$.

The system dynamics error on the $k^\mtext{th}$ grid partition is approximated by a basis function 
$\boldsymbol{\phi}( \xud{k+1} , \xud{k}, \uud{k}, \vardt_k )  = (\xud{k+1}\boxminus\xud{k})\vardt_k^{-1} - \boldsymbol{\xi}(\xud{k+1} , \xud{k}, \uud{k})$
with a suitable finite difference kernel $\boldsymbol{\xi}(\cdot)$,
 i.e. either forward differences (explicit, first order, similar to multiple shooting with forward Euler and full discretization):
\begin{equation}
\boldsymbol{\xi}(\xud{k+1} , \xud{k}, \uud{k}) \mdef \f(\xud{k}, \uud{k}), \label{eq:fwd_diff}
\end{equation}
or Crank-Nicolson differences (implicit, second order):
\begin{equation}
\boldsymbol{\xi}(\xud{k+1} , \xud{k}, \uud{k}) \mdef 0.5\big(\f(\xud{k}, \uud{k}) + \f(\xud{k+1}, \uud{k})\big).
\label{eq:crank_nic}
\end{equation}
By further approximating the integral cost in~\eqref{eq:ocp} by the right Riemann sum and $\ucdot{t}$ by forward differences, the resulting NLP is given as follows:
\begin{gather}
	\underset{ \substack{ \uud{0}, \uud{1}, \dotsc, \uud{N-1}, \\ \xud{0}, \xud{1}, \dotsc, \xud{N}, \\  \vardt_0, \vardt_1, \dotsc, \vardt_{N-1}} }{\min}\ \Jf\big(\xud{N}\big) + \sum_{k=0}^{N-1} \ell(\xud{k}, \uud{k}) \vardt_k
	\label{eq:local_uniform_nlp} \\
	\hspace{-6.8cm}\text{subject to} \nonumber \\ 
	\begin{align*}
		&\xud{0} = \xs, \ \xud{N} \in \xfset(\ts),\ \xud{k} \in \rxset(k \vardt_k), \ \uud{k} \in \ruset\big( \xud{k}, k \vardt_k \big),\\ 
		& (\uud{k+1}-\uud{k})\vardt_k^{-1} \in \rudset, \  (\uud{0}-\uud{\mtext{p}})\vardt_{\mtext{p}}^{-1} \in \rudset,\\
		&\vardtmin \leq \vardt_0 \leq \vardtmax,\quad \vardt_k = \vardt_{k+1},\\ 
		& \boldsymbol{\phi}( \xud{k+1} , \xud{k}, \uud{k}, \vardt_k ) = \nullvec,\quad k=0,1,\dotsc,N-1.
	\end{align*}
\end{gather}
The limitation of the first control w.r.t. its derivative at $t_0$ requires the specification of the previous control input $\uud{\mtext{p}} \in \uset$ and the time elapsed since $\uud{\mtext{p}}$, i.e. $\vardt_{\mtext{p}}=t_{n}-t_{n-1}$.
In the very first closed-loop step~\eqref{eq:mulaw}, $\uud{\mtext{p}}$ is set to zero or some control for which $\f(\xs, \uud{\mtext{p}})=\nullvec$ holds. The same holds for the last control $\uud{N}$ which is not subject to optimization but provides the 
final boundary value for the control derivative. This way, a robot with velocity input always plans to stop at the end of the horizon for safety reasons. 
Note that this is not conventional in receding-horizon \MPC{} as this inherently leads to differing open-loop and closed-loop solutions and might affect convergence to $\xfset(\ts)$.  
However, with terminal equality conditions as in the time-optimal variant shown later, this does not lead to changes in terms of convergence due to the optimality principle.
Condition $\tf \in I$ in~\eqref{eq:ocp} has been replaced by $\vardtmin \leq \vardt_0 \leq \vardtmax$ with bounds $\vardtmin,\vardtmax \in \realposzero$. 
Furthermore, condition $\vardt_k = \vardt_{k+1}$ ensures uniformity between all time intervals.
Notice that for the set evaluations $t_k = k\vardt_0 = k\vardt_k$ holds due to the uniform grid and choosing $k\vardt_k$ preserves the local structure of the optimization problem.

\begin{rem}
	NLP~\eqref{eq:local_uniform_nlp} is derived w.r.t. individual optimization parameters for each time interval but with constraints enforcing uniformity (local uniform grid approach). 
	Another formulation, the global uniform grid, is obtained by replacing all $\vardt_k$ with a single time parameter $\vardt$ and omitting constraints $\vardt_k = \vardt_{k+1}$.
	The optimal solution is identical, but the structure of the optimization problem differs slightly. For more details refer to~\cite{roesmann2020_stability}.
\end{rem}

\subsection{Receding-Horizon Quadratic-Form MPC} 
\label{sec:quad_form_mpc}

The most common \MPC{} realization is defined in terms of a quadratic-form objective to minimize the quadratic control error and effort.
To account for the alternative state space representations defined in \refsec{sec:alternative_ss}, the control error metric must be adjusted.
Common metrics for rotation groups are provided in~\cite{lavalle2006}, but in this case we include the nonlinear difference operator in the quadratic form.
The terminal and running costs are given as follows:
\begin{align}
\Jf(\xud{N}) &= (\xud{N} \boxminus \xf)^\transpose \Qf (\xud{N} \boxminus \xf), \label{eq:quad:jf} \\
 \ell(\xud{k}, \uud{k}) &= (\xud{k} \boxminus \xf)^\transpose \Q (\xud{k} \boxminus \xf) + \uud{k}^\transpose \R \uud{k}.  \label{eq:quad:ell}
\end{align}
Hereby, $\Q,\Qf \in \mathbb{R}^{p\times p}$ and $\R \in \mathbb{R}^{q\times q}$ denote weighting matrices with state dimension $p = \dim(\xset)$ and input dimension $q = \dim(\uset)$, respectively.
Furthermore, $\xf\in \xfset(t_n)$ represents the current intermediate goal state. 
In case the full goal state is not available, it is possible to either choose additional states to complete a steady state or to set related components in $\Q$ and $\Qf$ to zero.
Note that $\xf$ could also be replaced by a time-dependent reference $\xf(t)$ during optimization for tracking control. 
In this \MPC{} realization, the horizon length resp. final time~$\tf$ is fixed to $ I = \{ N \vardtref \}$ which implies $\vardtmin = \vardtmax = \vardt_k = \vardtref$ with a predefined grid resolution $\vardtref\in\realpos$.
Therefore the optimization w.r.t. time becomes obsolete and thus the optimization parameters $\vardt_k$ with $k=0,1,\dots,N-1$ as well as temporal constraints are excluded to reduce the dimensions of the optimization problem.

Stability and recursive feasibility conditions for \MPC{} usually rely on proper terminal conditions $\Jf(\cdot)$ resp.~$\Qf$ and $\xfset(t_n)$ to, e.g., approximate an infinite horizon.
The interested reader is referred to~\cite{gruene2017_book} and notice that a sampled-data discrete-time representation follows by setting $\vardtref=1$.
For (output) path-following MPC, \cite{faulwasser2016_tac} provides terminal conditions and conditions on the reference path under which stability is guaranteed. 
However, as a suitable terminal region is difficult to determine, most practical approaches usually relay on sufficiently long horizon lengths to ensure stability and feasibility and choose $\xfset(t_n)\mdef\xset$. 
For static environments and a class of robotic systems, i.e. differential drive robots, \cite{mehrez2017} guarantees stability for shorter receding horizons.  

\subsection{Time-Optimal MPC}
\label{sec:to_mpc}

Another interesting realization of NLP~\eqref{eq:local_uniform_nlp} is a pure time-optimal MPC as addressed for real spaces in~\cite{roesmann2020_stability}.
By setting $\Jf(\xud{N}) \mdef 0$ and $\ell(\xud{k}, \uud{k}) \mdef 1$ in \eqref{eq:local_uniform_nlp}, the resulting state trajectory reaches $\xfset(t_n)$ in minimum time.
$\vardtmax$ is set to a worst case accuracy for the system dynamics error or is even omitted ($\vardtmax = \infty$) as the grid adaptation scheme described below better controls the accuracy in changing environments.
For example, approaching obstacles extend the trajectory (longer traveling times for detours) while obstacles moving away contract the trajectory due to time-optimality. 
Similar as for the TEB approach~\cite{roesmann2017}, the temporal resolution $\vardt_k$ is adapted during closed-loop control w.r.t. a given reference $\vardtref \in \realpos$ and hysteresis $\vardthyst\in \realpos$:
	\begin{align*}
	N_{n+1} &=
	\begin{dcases}
		N_{n}+1 & \vardtstar_{n} > \vardtref + \vardthyst \\
		\max (N_{n}-1,\Nmin)    & \vardtstar_{n} < \vardtref - \vardthyst
	\end{dcases}.
\end{align*}
The grid size at closed-loop time $t_n$ is denoted by $N_n$ with initial size $N_0 = N \geq \Nmin$ and safe guard $\Nmin \in \naturalpos$.
$\vardtstar_{n}$ denotes the time interval obtained from the solution of NLP~\eqref{eq:local_uniform_nlp} at closed-loop time $t_n$.
Note that this linear search lowers the changes between subsequent solver calls (numerical robustness), but requires some iterations to react on highly dynamic environments.
In practice, changing the grid size one by one pointed out to be sufficient as the control rate is much faster than the changes in the environment. If this is not the case,
grid adaptation may change $N_n$ further or, alternatively,  an inner loop terminates NLP~\eqref{eq:local_uniform_nlp} prior to convergence and an outer loop adapts the grid size. 

For time-optimal control, the terminal condition is often set to the intermediate goal state $\xf$, i.e. $\xfset=\{ \xf \}$.
Closed-loop convergence and recursive feasibility results are provided in~\cite{roesmann2020_stability}.

\subsection{Obstacle Avoidance Constraints}

In this work, obstacles are defined as simply connected regions $\mathcal{O}_l(t), l =1,\dotsc, L$ with $L$ as the number of obstacles.
The time-dependency explicitly accounts for dynamic obstacles.
Furthermore, let $\mathcal{R}(\x)$ denote a geometric robot collision model dependent on the state $\x \in \xset$, e.g. position and orientation.
Usually these sets are embedded in \textit{SE(2)} or \textit{SE(3)}.
The minimum distance $\fundef{d_l}{\xset \times \mathbb{R}}{\realposzero}$  between the robot and obstacle~$l$ is then determined by:
\begin{equation}
	d_l(\x,t) = \min_{p_1 \in \mathcal{R}(\x), p_2 \in \mathcal{O}_l(t)} \ \lVert p_2 - p_1 \rVert_2.
\end{equation}
By specifying a minimum separation to obstacles $\dmin\in \realposzero$, the collision-free state space $\rxoset(t)$ is given as follows:
\begin{equation}
\rxoset(t) = \{ \x \in \xset \mid d_l(\x,t) \geq \dmin, l=1,2,\dotsc,L \}.
\end{equation}
Furthermore, let the robot's internal state constraints are described by set $\rxiset(t)\in \xset$, then $\rxset(t)$ in \eqref{eq:ocp} and \eqref{eq:local_uniform_nlp} is given by $\rxset(t) = \rxoset(t) \cap \rxiset(t)$.
Note that there exist techniques in the literature for computationally efficient optimization-based obstacle avoidance, e.g.~\cite{zhang2017optimizationbased}.

\subsection{Solution of the Nonlinear Program}
\label{sec:optim}

For NLP~\eqref{eq:local_uniform_nlp} general necessary and sufficient optimality conditions apply~\cite{nocedal2006_book}.
Any practical implementation replaces constraint sets $\rxset,\ruset,\rudset$ and $\xfset$ by algebraic equality and inequality functions. 
The NLP is solved by standard local constrained optimization techniques, such as interior point methods or sequential quadratic programming. 
For the time-optimal formulation even sequential linear programming can be applied. 
Warm-starting the NLPs in subsequent closed-loop steps with the previous solution significantly increases performance.
To comply with the alternative parameterizations (\refsec{sec:alternative_ss}), the solvers need to be modified in a few places.
Let $\mathbf{z} = (\mathbf{u}_0, \mathbf{x}_0, \vardt_0, \dotsc, \mathbf{u}_{N-1}, \mathbf{x}_{N-1}, \vardt_{N-1}, \mathbf{x}_N)$ define the optimization parameter vector of~\eqref{eq:local_uniform_nlp}.
Local solutions $\mathbf{z}^* = \mathbf{z} \boxplus \Delta \mathbf{z}$ after each solver step follow by incrementing components as follows: $\mathbf{u}_k + \Delta \mathbf{u}_k$, $\vardt_{k} + \Delta \vardt_k$ and \mbox{$\mathbf{x}_k \boxplus \Delta \mathbf{x}_k$}.
Also the calculation of Jacobians and gradients w.r.t. $\mathbf{x}$ requires to apply proper increments. E.g. the Jacobian of some vector valued state-dependent function $\mathbf{g}(\cdot)$ evaluated at $\bar{\mathbf{x}}$ is given by:
\begin{equation}
\jacobwrt{\mathbf{g}}{\mathbf{x}}(\bar{\mathbf{x}}) = \frac{\partial \mathbf{g}( \bar{\mathbf{x}} \boxplus \Delta \mathbf{x} ) }{ \partial\Delta \mathbf{x} } \bigg|_{\Delta \mathbf{x}=\nullvec} \label{eq:jacob_state}
\end{equation}
Since second-order derivatives can be calculated numerically using quasi-Newton methods or finite differences from first-order information, no dedicated increment is required.
\begin{rem}
Note that the nonlinear increment in~\eqref{eq:jacob_state} can be omitted for \textit{SO(2)} components in case $\mathbf{g}$ inherently normalizes the components. 
For example, consider the collocation constraint $\mathbf{g} \mdef \boldsymbol{\phi}(\cdot)$:
Here $\xud{k+1}\boxminus\xud{k}$ also applies to unnormalized components and the robot system dynamics $\f(\cdot)$ often considers rotational components only within trigonometric functions.
The same applies to~\eqref{eq:quad:jf},~\eqref{eq:quad:ell} and standard constraints including obstacle avoidance.
\end{rem}

The structure of NLP~\eqref{eq:local_uniform_nlp} is sparse and utilizing sparse algebra for efficient optimization is crucial.
While established NLP solvers like IPOPT~\cite{waechter2006_matprog} already support sparse algebra,
it is possible to further speed up optimization by computing derivative information only for structured non-zeros.
Similar to the TEB, NLP~\eqref{eq:local_uniform_nlp} is represented as a hypergraph with cost functions and constraints as hyperedges and states, controls and time invervals as vertices.
This facilitates the computation of sparse finite differences by iterating the edge set. 
A major advantage of the hypergraph is its simple algorithmic reconfiguration at negligible overhead.
This is crucial for real-time control as the NLP dimensions change during runtime either due to changing environments or grid adaptation.
Refer to~\cite{roesmann2018_aim} for general performance results and a detailed description on how to formulate NLPs in MPC as hypergraph.

 \subsection{Local Planning in Distinctive Topologies}
 \label{sec:topologies}

Solving NLP~\eqref{eq:local_uniform_nlp} as described before leads to locally optimal solutions.
Many local minima are caused by the presence of obstacles that imply non-convexity.
Identifying these local minima in advance coincides with exploring and analyzing distinctive topologies between start and goal poses.
Due to the limited scope of this paper, we cannot provide a detailed description, but would like to point out that distinctive topologies (homotopy classes) 
can be managed during runtime analogous to~\cite{roesmann2017} for $\xfset=\{\xf\}$. 
Several NLPs~\eqref{eq:local_uniform_nlp}  with state trajectories initialized for each active class are solved in parallel (multi-threaded). 
For dynamic obstacles and thus ($x$-$y$-$t$)  spaces, the equivalence relation that identifies those classes is changed to a 3D variant according to~\cite{bhattacharya2011}.
And for evaluating this relation, sampled roadmaps in~\cite{roesmann2017} are augmented with temporal profiles (maximum velocities). 


\section{EVALUATION}
\label{sec:evaluation}

The proposed MPC-based planning approach copes with arbitrary system dynamics \eqref{eq:system_dynamics_cont}  for states embedded in \textit{SE(2)} or any larger space containing \textit{SE(2)}/\textit{SO(2)} subsets.
Note that this includes common kinematic and dynamic models of mobile robots, vehicles and ships with velocity, acceleration, jerk, force and torque input spaces.
The following sections examine two common applications in more detail.

\subsection{Kinematic Bicycle Model}

For motion planning and control in autonomous driving, the kinematic bicycle model is often preferred over dynamic single or double track models due their sufficiently high accuracy over a large range of operation and less computational burden~\cite{kong2015}.
Let $x \in \mathbb{R}$ and $y \in \mathbb{R}$ define the coordinates of the center of mass and $\theta \in \textit{SO(2)}$ the inertial heading of the vehicle. 
Furthermore, $l_\mtext{f} \in \realposzero$ and $\l_\mtext{r} \in \realposzero$ are the distances from the center of mass to the front and rear axles, respectively.
Given a front steering angle $\delta \in (-\pi/2,\pi/2)$, the angle $\beta$ between the current velocity vector at the center of mass and the direction of the longitudinal axis of the vehicle is given by:
\begin{equation}
\beta(t) = \tan^{-1}\bigg( \frac{l_\mtext{r}}{l_\mtext{l}+l_\mtext{r}}  \tan\big( \delta(t) \big) \bigg).
\end{equation}
With state vector $\mathbf{x}(t)=\big(  x(t), y(t), \theta(t) \big)^\transpose$, vehicle speed $v(t)\in \mathbb{R}$ and control input $\mathbf{u}(t)=\big(  v(t), \delta(t) \big)^\transpose$, the kinematic bicycle model is defined by~\cite{kong2015}:
\begin{align}
\dot{\mathbf{x}}(t) = \f\big(\mathbf{x}(t), \mathbf{u}(t)\big) =
\begin{pmatrix}
 v(t) \cos\big( \theta(t) + \beta(t)\big) \\
 v(t) \sin\big( \theta(t) + \beta(t) \big) \\
 \frac{v(t)}{l_\mtext{r}} \sin\big(\beta(t)\big)
\end{pmatrix}.
\end{align}
Note, for this evaluation we choose speed $v$ as input rather than the acceleration $\dot{v}$ in contrast to~\cite{kong2015} and hence omit the additional integrator dynamics resp. state.
Limits on accelerations are still included by control deviation bounds.

The planning task constitutes an automated parking scenario, since the advantages of proper planning in \textit{SE(2)}/\textit{SO(2)} are particularly apparent here. 
When planning overtaking and (non-u-shape) turning maneuvers, $\theta$ usually does not reach $\pm \pi$ w.r.t. the local planning coordinate system.
Vehicle parameters $l_\mtext{f}=$\SI{1.1}{m} and $l_\mtext{r}=$\SI{1.7} are borrowed from~\cite{kong2015} for a Hyundai Azera.

\begin{figure}[tb]
	\centering
	%
	%
	\tikzset{draw/.append style={line width= 0.1mm}} 
	\begin{subfigure}[b]{\columnwidth}
		\centering
		\setlength{\figurewidth}{0.925\columnwidth}
		\setlength{\figureheight}{4cm}
		\includegraphics{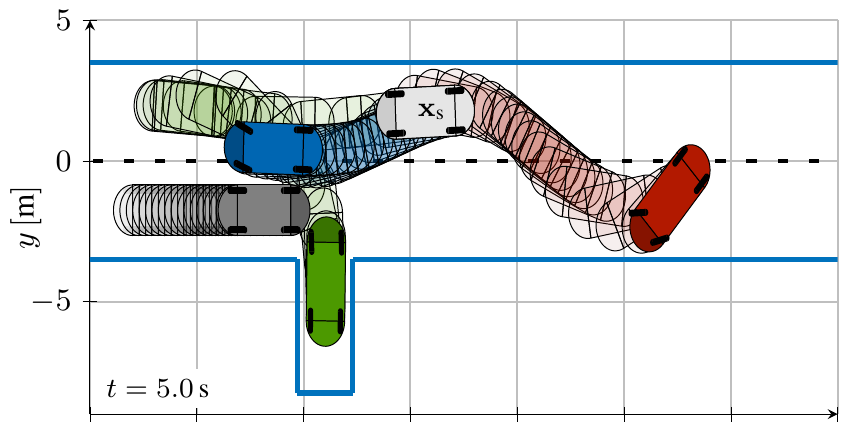}
	\end{subfigure} \\[-1.5ex]
	\begin{subfigure}[b]{\columnwidth}
		\centering
		\setlength{\figurewidth}{0.925\columnwidth}
		\setlength{\figureheight}{4cm}
		\includegraphics{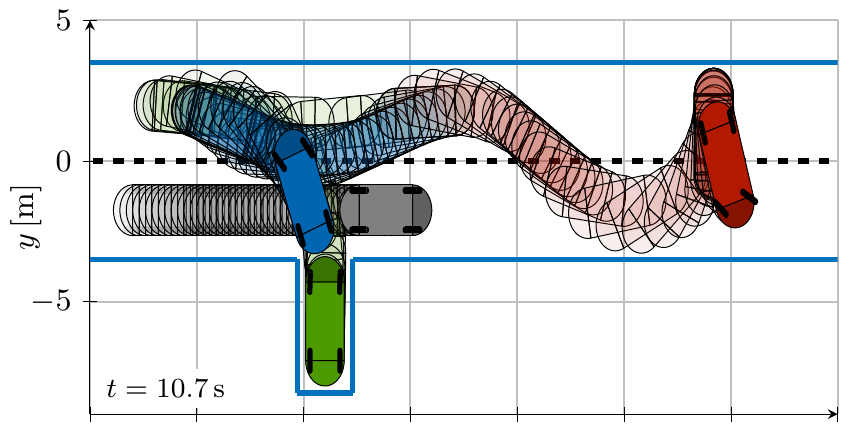}
	\end{subfigure} \\[-1.5ex]
	\begin{subfigure}[b]{\columnwidth} 
		\centering
		\setlength{\figurewidth}{0.925\columnwidth}
		\setlength{\figureheight}{4cm}
		\includegraphics{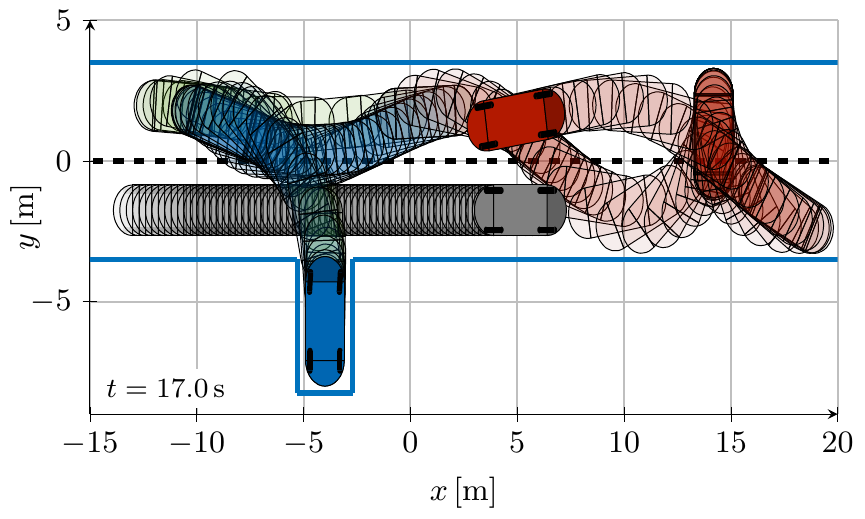}
	\end{subfigure}
	\caption{Quasi-time-optimal parking maneuver with dynamic obstacle (gray). Topologically distinctive solutions in \mbox{($x$-$y$-$t$)} are shown in blue and green. The red solution is obtained with standard Euclidean optimization. 
		The initial pose $\xs$ is shown in the top plot and the temporal resolution of traces is $0.3\,\mtext{s}$. The front part of each vehicle is filled with slightly darker color.}
	\label{fig:automatic_parking_xy}
\end{figure}

Control inputs are restricted by $\ruset$, i.e. $|v(t)|\leq 4\,\mtext{m/s}$ and $|\delta(t)|\leq 0.65\,\mtext{rad/s}$. 
Set $\rudset$ represents control deviation bounds ensuring $-3\,\mtext{m/s}^2 \leq \dot{v}(t) \leq 1.5\,\mtext{m/s}^2$ and $|\dot{\delta}(t)| \leq 0.31\,\mtext{rad/s}^2$. Internal states are not restricted, i.e. $\rxiset(t)\mdef\xset$.
The vehicle's start state is $\xs = (1\,\mtext{m}, 1.75\,\mtext{m}, -3.1\,\mtext{rad})^\transpose$ with $\mathbf{u}_\mtext{p} = \nullvec$ and its parking destination is $\xf = ( -4\,\mtext{m}, -6\,\mtext{m}, 1.57\,\mtext{rad})^\transpose$ with $\mathbf{u}_\mtext{N} = \nullvec$.
Road boundaries and the parking lot are modeled by six line segments according to~\reffig{fig:automatic_parking_xy} and define obstacle sets $\mathcal{O}_1,\mathcal{O}_2,\dotsc,\mathcal{O}_6$. The road width is \SI{3}{m} and the depth of the parking lot is \SI{4.75}{m}.
A pill resp. stadium-shape footprint with $w = l_\mtext{r}+l_\mtext{f}$ and $r = 0.9\,\mtext{m}$ serves as collision model:
\begin{equation}
\footnotesize
\thinmuskip 2.0mu
\medmuskip 0.5mu
\mathcal{R}(\mathbf{x}) = \Big\{ \mathbf{p} \in \mathbb{R}^2\, \Big\lvert\, \Big\lVert\!\! \begin{pmatrix} x+ \big(\mu  w -l_\mtext{r}  \big) \cos \theta  \\ y + \big(\mu w -{l}_\mtext{r}  \big) \sin \theta  \end{pmatrix} - \mathbf{p} \Big\rVert_2 \leq  r\, \forall \mu \in [0,1] \Big\}.
\nonumber
\end{equation}

To demonstrate that the approach is also suitable for dynamic obstacles, although the time during optimization can be variable, the following obstacle with constant speed $v_\mtext{o} = 1\, \mtext{m/s}$ is added.
The start position is on the other lane at $x_\mtext{o} = -13\,\mtext{m}$ and $y_\mtext{o} = -1.25\,\mtext{m}$. 
Its time-dependent collision model with $r_\mtext{o}=0.9\,\mtext{m}$ and $l_\mtext{o}=2.5\,\mtext{m}$ is given as follows:
\begin{equation}
\footnotesize
\mathcal{O}_7(t) = \Big\{ \mathbf{p} \in \mathbb{R}^2\, \Big\lvert\, \Big\lVert \begin{pmatrix} x_\mtext{o}+ \mu  l_\mtext{o} + t v_\mtext{o}   \\ y_\mtext{o}  \end{pmatrix} - \mathbf{p} \Big\rVert_2 \leq  r_\mtext{o}\, \forall \mu \in [0,1] \Big\}.
\nonumber
\end{equation}
The minimum separation is set to $\dmin = 0.2\,\mtext{m}$.
We choose a variable time realization according to \refsec{sec:to_mpc}. 
We merely extend the running costs by a small weighted term for the control effort, making it a hybrid objective: $\ell(\mathbf{x}_k,\mathbf{u}_k) = 1 + \mathbf{u}_k^\transpose \mathbf{R} \mathbf{u}_k$ with $\mathbf{R}=\operatorname{diag}(0.01,0)$.
The low weight still leads to quasi-time-optimal solutions, but the vehicle tends to prefer braking to swerving out, especially if it is waiting for the dynamic obstacle to pass.
Local planning is based on Crank-Nicolson differences~\eqref{eq:crank_nic} and includes the final goal state as terminal condition, i.e. $\xfset = \{ \xf \}$.
As the state space is the \textit{SE(2)}, operator overloading follows from \refexample{ex:se2}.
Further parameters are $N=50$, $\vardtref=\vardt_\mtext{p}=0.1\,\mtext{s}, \vardtmin=0.001\,\mtext{s}$, $\vardtmax=\infty$, $\Nmin=2$, $\vardthyst=0.1\vardtref$.

\begin{figure}[tb]
	\centering
%
%
\definecolor{mycolor1}{rgb}{0.00000,0.40000,0.70000}%
\definecolor{mycolor2}{rgb}{0.30000,0.60000,0.00000}%
\definecolor{mycolor3}{rgb}{0.70000,0.10000,0.00000}%
\begin{tikzpicture}

\setlength{\figurewidth}{0.85\columnwidth}
\setlength{\figureheight}{1.4cm}

\tikzstyle{style1}=[mycolor1,solid,line width=1.25pt]
\tikzstyle{style2}=[mycolor2,solid,line width=1.25pt] 
\tikzstyle{style3}=[mycolor3,solid,line width=1.25pt] 

\begin{axis}[%
width=\figurewidth,
height=\figureheight,
name=plotv,
at={(0\figurewidth,0.581\figureheight)},
scale only axis,
xmin=0.000,
xmax=22,
extra x ticks={22},
extra x tick labels={},
xticklabels={,,}, 
ymin=-4.000,
ymax=4.000,
ylabel={$v\, [\text{m}/\text{s}]$},
y label style={at={(-0.11,0.5)}},
axis background/.style={fill=white},
axis x line*=bottom,
axis y line*=left,
xmajorgrids,
ymajorgrids,
legend columns = 3,
legend pos = outer north,
legend style={draw=none,fill=none,legend cell align=left, column sep =4pt, font=\scriptsize,at={(\figurewidth/2-0.3cm,\figureheight+0cm)}},
]
\addplot[ style1, forget plot] table[] {automatic_parking_controls-1.tsv};
\addplot[ style2, forget plot] table[] {automatic_parking_controls-2.tsv};
\addplot[ style3, forget plot] table[] {automatic_parking_controls-3.tsv};

\addlegendimage{style1};
\addlegendentry{HC1};
\addlegendimage{style2};
\addlegendentry{HC2};
\addlegendimage{style3};
\addlegendentry{No \textit{SE(2)}-aware Planning};
\end{axis}

\begin{axis}[%
width=\figurewidth,
height=\figureheight,
name=plotdelta,
at=(plotv.below south west),
anchor=above north west,
yshift=0.5\baselineskip,
scale only axis,
xmin=0.000,
xmax=22,
extra x ticks={22},
extra x tick labels={},
xticklabels={,,}, 
ymin=-0.800,
ymax=0.800,
extra y ticks={0.8},
extra y tick labels={},
ylabel={$\delta\, [\text{rad}]$},
y label style={at={(-0.11,0.5)}},
axis background/.style={fill=white},
axis x line*=bottom,
axis y line*=left,
xmajorgrids,
ymajorgrids
]
\addplot[ style1, forget plot] table[] {automatic_parking_controls-4.tsv};
\addplot[ style2, forget plot] table[] {automatic_parking_controls-5.tsv};
\addplot[ style3, forget plot] table[] {automatic_parking_controls-6.tsv};
\end{axis}

\begin{axis}[%
width=\figurewidth,
height=\figureheight,
name=plottheta,
at=(plotdelta.below south west),
anchor=above north west,
yshift=0.5\baselineskip,
scale only axis,
xmin=0,
xmax=22,
extra x ticks={22},
extra x tick labels={},
xlabel={$t\,[\text{s}]$},
ymin=-4.000,
ymax=4.000,
y label style={at={(-0.11,0.5)}},
ylabel={$\theta\,[\text{rad}]$},
axis background/.style={fill=white},
axis x line*=bottom,
axis y line*=left,
xmajorgrids,
ymajorgrids
]
\addplot [style1, forget plot] table[]{automatic_parking_controls-7.tsv};
\addplot [style2, forget plot] table[]{automatic_parking_controls-8.tsv};
\addplot [style3, forget plot] table[]{automatic_parking_controls-9.tsv};
\end{axis}
\end{tikzpicture}%
	\caption{Control input and orientation profiles for the parking scenario in~\reffig{fig:automatic_parking_xy}}
	\label{fig:automatic_parking_controls}
\end{figure}
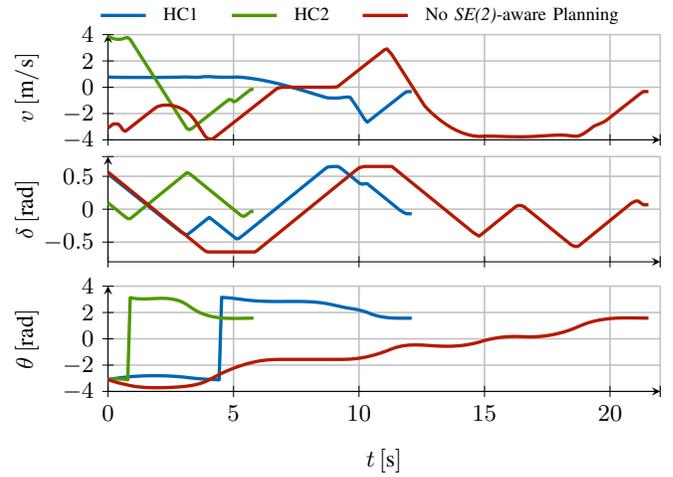

\reffigc{fig:automatic_parking_xy} shows the open-loop solutions obtained after convergence for two topologically distinctive initializations resp. homotopy classes (HC) in \mbox{($x$-$y$-$t$)} space. 
The related control input and orientation profiles are depicted in \reffig{fig:automatic_parking_controls}.
While in the first HC the ego vehicle slows down and waits for the obstacle to pass (blue), in the second HC it reaches the parking lot before the obstacle passes (green).
Note, the variable time horizon is very advantageous here to find feasible trajectories that ultimately reach $\xf$.
We would like to point out that the chosen obstacle distances are deliberately kept small in order to demonstrate the abilities and limit mathematical exposition.
Adding further potential fields to the running cost and preferring control effort to time optimality increases safety and comfort in practice.
The third solution (red) follows from solving \eqref{eq:local_uniform_nlp} without proper nonlinear operators for \textit{SO(2)} components. 
The orientation $\theta(t)$ continuously rotates from $-3.1\,\mtext{rad}$ to $1.57\,\mtext{rad}$. 
In contrast, $\theta(t)$ in HC1 and HC2 jumps without affecting the local continuity of the optimization problem itself.
Note that it would also be possible to get the desired behavior in this scenario by adjusting the start and goal orientations by multiples of $2\pi$. 
However, this additional logic is error-prone and not as flexible and complete as the explicit consideration in the optimization. 

\subsection{Comparative Analysis for a Differential Drive Robot}

The parking scenario with the bicycle model shows how versatile the approach is.
The TEB approach also supports car-like robots, but only with a simple model and without steering rate limits, so the previous example would not have been feasible.
This makes our approach suitable for especially more complex models and scenarios. 
In this section, we conduct a comparative analysis for differential drive robots in a common hierarchical planning framework. 
Simulative experiments are carried out with the navigation stack in ROS, for which several local planners for differential drive robots exist,
e.g. the classic EB, DWA resp. trajectory rollout, TEB, and our new MPC-based planner\footnote{http://wiki.ros.org/\{eband,dwa,teb,mpc\}\_local\_planner}.
The simulation environment is \textit{stage} in ROS which runs in a separate process and thus also simulates communication and calculation delays.

\begin{figure}[tb]
	\centering
	\input{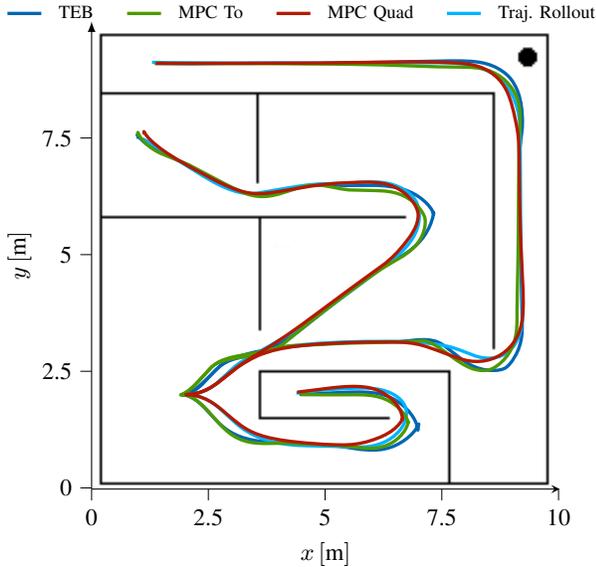}
	\caption{Navigation of a differential drive robot for three different goals and several methods}
	\label{fig:diff_drive_map}
	\vspace{-0.5\baselineskip}
\end{figure}

The circular robot with collision model
\begin{equation}
\mathcal{R}(\mathbf{x}) = \big\{ \mathbf{p} \in \mathbb{R}^2 \mid \lVert (x,y)^\transpose - \mathbf{p} \rVert_2 \leq 0.17\,\mtext{m} \big\}
\end{equation}
is initially located at $(2\,\mtext{m},2\,\mtext{m},0\,\mtext{rad})^\transpose$.
The navigation task is to drive to three goals located at $(1\,\mtext{m},7.5\,\mtext{m},\pi/2\,\mtext{rad})^\transpose$, $(1.2\,\mtext{m},9.1\,\mtext{m},3.14\,\mtext{rad})^\transpose$ and $(4.3\,\mtext{m},2.0\,\mtext{m},-3.14\,\mtext{rad})^\transpose$  in a $10\,\mtext{m} \cdot 10\,\mtext{m}$ map as shown in \reffig{fig:diff_drive_map}. The kinematic model in \textit{SE(2)} and linear and angular velocities as input $\mathbf{u}=(v,\omega)^\transpose$ is given by:
\begin{align}
	\f\big(\mathbf{x}(t),\mathbf{u}(t)\big)=
	\begin{pmatrix}
		v(t) \cos\big( \theta(t)\big),
		v(t) \sin\big( \theta(t) \big) ,
		\omega(t)
	\end{pmatrix}^\transpose. \nonumber
\end{align}
Considered limits are $-0.2\,\mtext{m/s} \leq v(t) \leq 0.4\,\mtext{m/s}$, $| \omega(t) | \leq 0.4\,\mtext{rad/s}$, $|\dot{v}(t)| \leq 0.25\,\mtext{m/s}^2$ and $|\dot{\omega}(t)| \leq 0.25\,\mtext{m/s}^2$.
Hierarchical planning is realized by the standard Dijkstra-based global planner in ROS which refreshes its path every \SI{2}{s}.
The local planners operate at \SI{10}{Hz} in a local costmap of size $5\,\mtext{m} \cdot 5\,\mtext{m}$ and a resolution of $0.1\,\mtext{m/cell}$ centered at the current robot location.
Intermediate goals for local planning, i.e. $\xf$, are centered on the global path within the local costmap and a lookahead distance of $1.5\,\mtext{m}$.
The planner parameters are almost set to the default setting in addition to those that meet the above mentioned kinodynamic constraints correctly.

Note that unlike the other planners, the classic EB is a pure path planning method, so the package also implements a PID-based path following controller. 
Modified parameters are the minimum bubble overlap ($0.2\,\mtext{m}$), costmap weight ($20$) and the velocity multiplier ($8$).

The DWA planner is configured in the computationally more expensive trajectory rollout mode, because the dynamic window does not work well with small acceleration limits like here and does not always converge.
Besides default parameters, the horizon length is set to $2\,\mtext{s}$, the granularity to $0.1\,\mtext{s}$ and the number of $v$- and $\omega$-samples to $10$ and $20$, respectively.
Note that this setting is not tuned for computational efficiency, but for comparable grid sizes. 

For TEB and MPC, each occupied cell is treated as a single point obstacle. The number of occupied cells in all three navigation runs is $103\pm27$. Prior to each optimization step, further filtering associates each discrete state only with nearby obstacles on the left and right side.
Three different configurations are considered for MPC:
\begin{compactenum}[i)]
 \item  A quadratic form realization according to \refsec{sec:quad_form_mpc} with $\Q=\Qf=\operatorname{diag}(1, 1, 0.25)$, $\R=\operatorname{diag}(2,2)$, $\xfset = \rxset$, $\vardtref=0.3\,\mtext{s}$ and $N=30$ (\textit{MPC Quad}),
\item  A minimum-time realization according to \refsec{sec:to_mpc} with $\xfset=\{\xf\}$ (\textit{MPC To}), \label{item2}
 \item A hybrid realization based on \ref{item2}) with $\ell(\cdot) = 1 + \mathbf{u}_k^\transpose \mathbf{R} \mathbf{u}_k$ and $\mathbf{R}=\operatorname{diag}(2,2)$ (\textit{MPC Hybrid}).
\end{compactenum}
The NLPs with forward differences~\eqref{eq:fwd_diff} are warm-started and solved with the established C++ interior point solver IPOPT~\cite{waechter2006_matprog} and sparsity exploitation according to \refsec{sec:optim}. 

\reftabc{tab:dif_drive_results} shows the benchmark results obtained with a PC running Ubuntu 18.04 (Intel Core i7-4770 CPU at \SI{3.4}{GHz}, \SI{8}{GB} RAM).
Travel time, path length and control effort $\big(\!\int_0^\infty \mathbf{u}(t)^\transpose\mathbf{u}(t)\,\dt\big)$ are the accumulated absolute values of all three movements from start to the three goals.
These values are comparable despite the classic EB. It is important to note that each approach can be further configured according to individual needs.
All approaches successfully reach their goals. Trajectory rollout and quadratic form MPC reveal slightly faster travel times, but this is due to the non existing terminal conditions and therefore prefer cutting corners (cf. \reffig{fig:diff_drive_map}).
More significant is the comparison of the CPU times, which are represented by their median and [0.05, 0.95]-quantiles.
The TEB approach requires the least computational resources but also the MPC solutions are obtained in a relatively short time, enabling their application in real-world scenarios at common controller rates.

\setlength{\tabcolsep}{1.25pt}
\renewcommand{\arraystretch}{0.96}
\begin{table}[tb]
	\caption{Benchmark results for different planners}
	\label{tab:dif_drive_results}
	\begin{tabularx}{\columnwidth}{p{1.7cm} Y Y Y Y}
		\toprule
		& CPU Time [ms] & Travel Time [s] & Path Length [m] & Control Effort \\ 
		\midrule 
		Elastic Band & \mbox{$10.2$ [$1.7$, $18.3$]} & $192.5$ & $48.1$ & $43.3$ \\
		Traj. Rollout & \mbox{$19.1$ [$14.8$, $22.1$]} & $122.1$ & $42.1$ & $20.9$ \\
		TEB & \mbox{$5.9$ [$3.0$, $9.4$]} & $126.9$ & $44.2$ & $21.9$ \\
		MPC To & \mbox{$12.5$ [$5.4$, $21.8$]} & $124.6$ & $43.1$ & $22.3$ \\
		MPC Quad & \mbox{$15.8$ [$7.5$, $25.3$]} & $116.0$ & $41.6$ & $20.4$ \\
		MPC Hybrid & \mbox{$14.2$ [$7.6$, $22.4$]} & $140.1$ & $41.9$ & $17.5$ \\
		\bottomrule
	\end{tabularx}
\end{table}

\section{CONCLUSIONS}
\label{sec:conclusions}

MPC is a powerful and highly customizable approach to robot motion planning. 
While approaches in the literature are usually based on purely Euclidean optimization spaces, 
the proposed nonlinear operator technique for the explicit treatment of rotational components in \textit{SO(2)} during optimization points out to be crucial for holistic, generic and cross-scenario motion planning.
Even though the CPU times for simple robotic applications (e.g. differential drive) are within the usual range, there are no significant advantages from a performance point of view, so that efficient planners such as the TEB approach need not be replaced.
But as soon as more complex kinematic or dynamic models or environments become necessary, 
the previous planners (at least in ROS) reach their limits. Our provided open-source C++ MPC framework with ROS integration is highly modular, versatile and computationally efficient.

\addtolength{\textheight}{0cm}   






\printbibliography

\end{document}